\documentclass{article} \usepackage{iclr2026_conference,times}

\usepackage{amsmath,amsfonts,bm}

\def\eqref#1{equation~\ref{#1}}

\def\1{\bm{1}}

\DeclareMathAlphabet{\mathsfit}{\encodingdefault}{\sfdefault}{m}{sl}
\SetMathAlphabet{\mathsfit}{bold}{\encodingdefault}{\sfdefault}{bx}{n}

\usepackage{hyperref}
\usepackage{url}
\usepackage{graphicx}
\usepackage{booktabs}

\usepackage[nohyperlinks]{acronym}
\acrodef{SSL}[SSL]{self-supervised learning}
\acrodef{AI}[AI]{artificial intelligence}
\acrodef{JEPA}[JEPA]{joint-embedding predictive architecture}
\acrodef{RL}[RL]{reinforcement learning}
\acrodef{MBRL}[MBRL]{model-based reinforcement learning}
\acrodef{EMA}[EMA]{exponential moving average}
\acrodef{ELBO}[ELBO]{evidence lower bound}
\acrodef{KL}[KL]{Kullback-Leibler}
\acrodef{WM}[WM]{world model}
\acrodef{CDP}[CDP]{continuous deterministic representation prediction}
\acrodef{CNN}[CNN]{Convolutional Neural Network}
\acrodef{MLP}[MLP]{Multilayer Perceptron}
\acrodef{RNN}[RNN]{recurrent neural network}
\acrodef{RSSM}[RSSM]{Recurrent State-Space Model}

\newcommand{\ours}{Dreamer-CDP}

\title{\ours{}: Improving Reconstruction-free World Models Via Continuous Deterministic Representation Prediction}

\author{Michael Hauri \& Friedemann Zenke \\
Friedrich Miescher Instiute for Biomedical Research\\
Fabrikstrasse 24\\
4056 Basel, Switzerland \\
\texttt{\{michael.hauri,friedemann.zenke\}@fmi.ch} \\
}

\iclrfinalcopy 
\begin{document}

\maketitle

\begin{abstract}
\Ac{MBRL} agents operating in high-dimensional observation spaces, such as Dreamer, 
rely on learning abstract representations for effective planning and control. 
Existing approaches typically employ reconstruction-based objectives in the observation space, which can render representations sensitive to task-irrelevant details. 
Recent alternatives trade reconstruction for auxiliary action prediction heads or view augmentation strategies, but perform worse in the Crafter environment than reconstruction-based methods. 
We close this gap between Dreamer and reconstruction-free models by introducing a JEPA-style predictor defined on continuous, deterministic representations. 
Our method matches Dreamer’s performance on Crafter, demonstrating effective world model learning on this benchmark without reconstruction objectives.
\end{abstract}

\section{Introduction}
Recent progress in \ac{MBRL} has enabled data-efficient learning in high-dimensional observation spaces.
Central to these methods is learning a latent dynamics (or ``world'') model, which is subsequently used for downstream tasks such as planning, control, and policy optimization. 
To be effective, this world model must operate on abstract, compressed representations that capture task-relevant structure while discarding irrelevant details.

\Ac{SSL} has proven effective for learning such representations from experience.  
Classic works rely on reconstruction objectives \citep{ha2018,hafner2019}.
However, reconstruction may bias representations toward pixel-level details that are largely irrelevant for behavior \citep{nguyen2021, zhang2024,voelcker2024}. 
This insight is generating interest in reconstruction-free \ac{SSL}.
To that end, several works studied reconstruction-free variants \citep{Deng2022, Burchi2024} of Dreamer, a widely-used \ac{MBRL} framework \citep{hafner2019b}. 
However, these strategies do not match the performance of reconstruction-based approaches on challenging benchmarks such as Crafter. 
This shortcoming may be a consequence of training both the representation and transition models to predict Dreamer's discrete probabilistic state variables.

We combine recent \ac{SSL} ideas into \ours{}, which learns a  world model by adding \ac{CDP}, while matching Dreamer's performance on Crafter.

\section{Prior Work}
\label{Sec:priorwork}
\textbf{Preventing collapse in reconstruction-free \ac{SSL}:} One approach is the use of contrastive objective functions  \citep{oord2018}.
However, contrastive methods often require large batch sizes \citep{chen2022}, typically violate temporal locality, and can suffer from the curse of dimensionality \citep{lecun2022}.  
Another strategy to avoid collapse leverages \acp{JEPA} \citep{lecun2022, garrido2024}, which trades contrastive objectives for additional regularization techniques \citep{bardes2021}, or predictor networks with specific stop-gradient placement such as BYOL \citep{grill2020} and SimSiam \citep{chen2021}.

\textbf{Reconstruction-free \ac{SSL} in \ac{RL}} has been explored in settings with high-dimensional input spaces, such as images. \citet{zheng2023b,zheng2023, burchi2025} used contrastive learning methods. Several other works employed self-predictive learning  \citep{ni2024} either to learn directly state representations \citep{gelada2019, schwarzer2020} or to train a world model for downstream control and planning tasks \citep{zhou2024, sobal2025, assran2025}.  
Closely related to our work, BYOL-explore \citep{guo2022} introduced a parsimonious design based on \citet{grill2020} to solve Atari games. In addition, EfficientZero \citep{ye2021,wang2024} and 
TD-MPC2 \citep{hansen2023} adopted a SimSiam-style architecture. However, in contrast to Dreamer models, these algorithms rely on a purely non-stochastic continuous-variable world model.   

Within the context of Dreamer, \citet{okada2022dreamingv2} proposed to use contrastive-learning methods, whereas \citet{Deng2022} integrated prototypical representations \citep{caron2020} to temporal dynamics learning in DreamerPro. In MuDreamer, \citet{Burchi2024} proposed to use the action signal to train the world model (Table~\ref{tab:components}). 
Despite this diversity of existing approaches, reconstruction-based models remain the gold standard for Crafter (see Table~\ref{tab:results}).  

\begin{table}[tbh]
    \centering
    \small
    \begin{tabular}{lcccccc}
        \toprule
        Method & Reconstruction-free & Non-contrastive & No action prediction& No view augmentation \\
        \midrule
        Dreamer     & $\circ$ & $\circ$ & $\bullet$ & $\bullet$ \\
        DreamerPro  & $\bullet$ & $\bullet$ & $\bullet$ & $\circ$ \\
        MuDreamer   & $\bullet$ & $\circ$ & $\circ$ &  $\bullet$ \\
        \ours{}  & $\bullet$ & $\bullet$ & $\bullet$ & $\bullet$ \\
        \bottomrule
    \end{tabular}
    \caption{Overview of Dreamer variants. Dreamer relies on pixel-based reconstruction. Other methods are reconstruction-free: 
    MuDreamer uses action prediction to train the sequence model. DreamerPro utilizes augmented views. 
    \ours{} (this article) relies solely on internal prediction.}
    \label{tab:components}
\end{table}

\section{The Dreamer framework}
\label{dreamer_framework}
Dreamer is an \ac{MBRL} algorithm that learns a world model and uses imagined trajectories for policy learning, thereby instantiating the Dyna framework \citep{sutton1991} in a high-dimensional, pixel-based setting. 
Here, we briefly recap the DreamerV3 implementation \citep{Hafner2025}.
The current observation $x_t$ is encoded into a discrete stochastic state $z_t$. The sequence model predicts the next hidden state $h_{t+1}$, from $h_t$, $z_t$, and the action leading to the next state $a_{t}$. The dynamics are learned by reconstructing the next input $\hat{x}_{t+1}$. The model is summarized below:
\begin{eqnarray*}
    \text{Sequence model:}~ h_{t} &= & f_\phi(h_{t-1}, z_{t-1}, a_{t-1}) \nonumber  \\
    \text{Encoder:}~ z_t & \sim & q_\phi(z_t|h_t,x_t) \nonumber \\
    \text{Dynamics predictor:}~ \hat{z}_t & \sim & p_\phi(\hat{z}_t|h_t) \nonumber \\
    \text{Reward predictor:}~ \hat{r}_t & \sim  &p_\phi(\hat{r}_t|h_t,z_t) \nonumber \\
    \text{Continuation flag predictor:}~ \hat{c}_t &  \sim  &p_\phi(\hat{c}_t|h_t,z_t) \nonumber \\
    \text{Decoder: }~ \hat{x}_t & \sim & p_\phi(\hat{x}_t|h_t,z_t)   
\end{eqnarray*}
where $c_t$ is the continuation flag and $r_t$ is the reward at time $t$. 
The world model is trained with the following loss: 
\begin{equation}
\mathcal{L}(\phi) = E_{q_\phi}\left[\sum_t\left(\beta_\mathrm{recon}\mathcal{L}_\mathrm{recon}(\phi)+\beta_\mathrm{aux}\mathcal{L}_\mathrm{aux}(\phi) + \beta_\mathrm{dyn}\mathcal{L}_\mathrm{dyn}(\phi)+ \beta_\mathrm{rep}\mathcal{L}_\mathrm{rep}(\phi)\right)\right]    
\end{equation}

with 
\begin{eqnarray*}
\arraycolsep=1.4pt\def\arraystretch{1.6}
\mathcal{L}_{\mathrm{recon}}(\phi) &=& -\ln p_\phi(x_t|z_t,h_t) \nonumber \\
    \mathcal{L}_{\mathrm{aux}}(\phi) &=& - \ln p_\phi(r_t|z_t,h_t) -\ln p_\phi(c_t|z_t,h_t) \nonumber  \\
    \mathcal{L}_{\mathrm{dyn}}(\phi) &=& \max(1,\text{D}_\mathrm{KL}[\mathrm{SG}(q_\phi(z_t|h_t,x_t))|| p_\phi(z_t|h_t)]) \nonumber \\
    \mathcal{L}_{\mathrm{rep}}(\phi) &=& \max(1,\text{D}_\mathrm{KL}[q_\phi(z_t|h_t,x_t)|| \mathrm{SG}(p_\phi(z_t|h_t))])  
\quad ,
\end{eqnarray*}
where SG is the stop-grad operator and $\mathrm{D_{KL}}$ is the \ac{KL} divergence. 
\section{Reconstruction-free World Model Learning in \ours}
\begin{figure}[tb]
    \centering
    \includegraphics[width=0.9\linewidth]{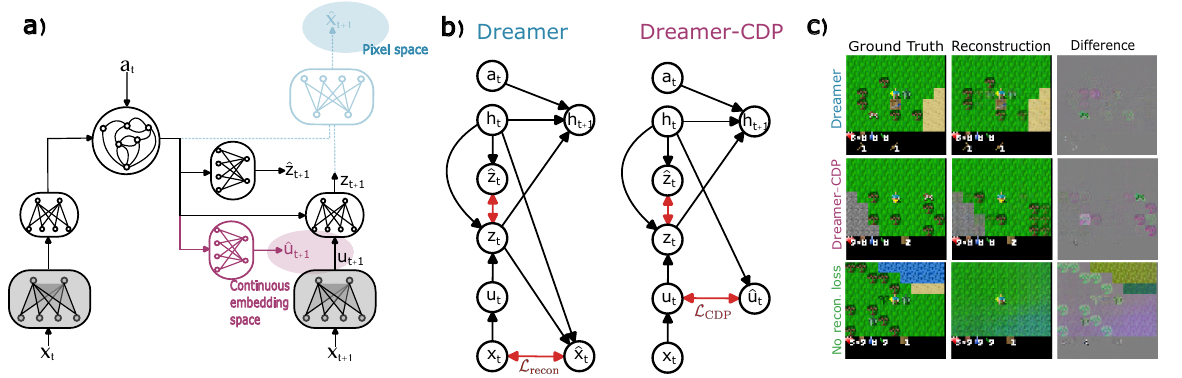}
    \caption{\textbf{a)}~Schematic of \ours{}. The hidden state is passed through a predictor (green) trained to approximate the next continuous representation $\hat{u}_{t} \approx u_{t}$.  In Dreamer, the hidden state and the input embedding are used to predict the next input $x_{t}$ (dashed gray).   \textbf{b)}~Graphical model of Dreamer (left) and \ours{} (right) with losses in red.  \textbf{c)}~Visual examples when $\mathcal{L}_\mathrm{recon}$ (Dreamer), $\mathcal{L}_\mathrm{CDP}$ (\ours{}) or neither is applied. For the latter two, the decoder was trained independently with detached gradients for visualization purposes.} 
    \label{fig:rpl}
\end{figure}

To show how to learn efficient world models without reconstruction, we introduce \ours{}\footnote{Code is available at: https://github.com/fmi-basel/Dreamer-CDP}, a simple variant of DreamerV3 \citep{Hafner2025} in which we remove the reconstruction loss while adding a JEPA-style predictor for \ac{CDP} (Fig.~\ref{fig:rpl}a,b) inspired by recent work on temporal \acp{JEPA} \citep{mohammadi2025}. 
To that end, we separate Dreamer's original representation encoder $q_\phi(z_{t}|h_{t},x_{t})$ as follows: 
First, observations $x_{t}$ are mapped to a continuous deterministic embedding $u_{t}$ via a feature extractor. A stochastic encoder then predicts a latent state representation $z_{t} \sim p_\phi(z_{t}|h_{t},u_{t})$ from the features $u_{t}$ and the hidden state $h_{t}$. The latent state, together with the current action $a_{t}$, is processed by a recurrent dynamics model to yield $h_{t+1}$. 
While a feedforward architecture would be sufficient under full observability \citep{hansen2023}, 
here we follow the Dreamer lineage in which the \ac{RNN} is essential to deal with
partial observability.
However, \acp{RNN} create distinct challenges for predictive \ac{SSL} \citep{mohammadi2025}.
To navigate these challenges, we train the predictor $\hat{u}_{t+1} = g_\phi(h_{t+1})$ on the continuous deterministic embeddings $u_{t+1}$ of the future observation $x_{t+1}$, which does not depend on the hidden state $h_{t+1}$. 
In contrast to \citet{guo2022}, we do not use an \ac{EMA} target network. 
Instead, we rely on the insight that the sequence model must be close to a fixed point of its dynamics when the parameters of the representation network are updated \citep{Tang2023,khetarpal2025}.
To ensure this convergence, we train the sequence model predictors with a higher learning rate (see \ref{appendix:hyperparams}) using
the following objective:
\begin{equation}
\mathcal{L}(\phi) = E_{q_\phi}\left[\sum_t\left(\beta_\mathrm{CDP}\mathcal{L}_\mathrm{CDP}(\phi)+\beta_\mathrm{aux}\mathcal{L}_\mathrm{aux}(\phi) + \beta_\mathrm{dyn}\mathcal{L}_\mathrm{dyn}(\phi)+ \beta_\mathrm{rep}\mathcal{L}_\mathrm{rep}(\phi)\right)\right]    
\end{equation}
where $\mathcal{L}_\mathrm{CDP}$ is given by the negative cosine similarity
$\mathcal{L}_\mathrm{CDP}(\phi) = - \sum_t \cos(\mathrm{SG}(u_{t+1}),\hat{u}_{t+1})$.

It is worth noting that the original Dreamer architecture already incorporates internal prediction through its \ac{KL} balancing mechanisms \citep{hafner2021}, which arises from the \ac{KL} regularization term in the \ac{ELBO} objective \citep{hafner2019}.
However, this prediction mechanism leverages the purely probabilistic discrete targets for learning (cf.\ Fig.~\ref{fig:rpl}b) and by itself does not lead to high-performing world models, as we will see below. 

\section{Experiments}
To evaluate \ours{}, we used Crafter \citep{hafner2021}, a computationally lightweight version of Minecraft, allowing us to assess agents on long-term reasoning, exploration, generalization, and dealing with sparse rewards. 
Performance was measured using the Crafter score, a metric that weighs the discovery of new achievements more strongly than the exploitation of already unlocked ones. For instance, unlocking a new achievement that is reached in only 1\% of episodes yields a larger score increase than improving the success rate of an existing achievement from 90\% to 100\%.

\textbf{Implementation.}
We trained all models on a single Nvidia V100 GPU. The model comprised a \acs{MLP} deterministic predictor and a \ac{RSSM} \citep{hafner2019} with a \acs{CNN} encoder  (see \ref{appendix:hyperparams} for details). Each model interacted 1M times with the environment. 

\textbf{Baseline methods.}
\label{baseline_methods}
We compared \ours{} to three different algorithms (Table~\ref{tab:components}). First, the original DreamerV3 \citep{Hafner2025}, which learns the world model by reconstruction in the input space (Sec.~\ref{dreamer_framework}). We also compared it to MuDreamer \citep{Burchi2024}, which, inspired by MuZero \citep{schrittwieser2020}, trains the world model by predicting the action $\hat{a}_t \sim p_\phi(\hat{a}_t|h_{t},z_{t}, x_{t+1})$ that lead to the current state and the value $\hat{v}_t \sim p_\phi(\hat{v}_t|h_t,z_t)$. MuDreamer attained performance comparable to Dreamer on the Atari and DeepMind Control benchmark, even outperforming it when trained with naturalistic backgrounds. Finally, we also compared it to DreamerPro \citep{Deng2022}, another non-contrastive \ac{SSL} method, which combines prototypical representations \citep{caron2020} with learning a sequence model, but by predicting jointly the cluster assignment of the observation and an augmented view rather than predicting the next latent state.

\subsection{Results}

\ours{} achieved a Crafter score of $16.2 \pm 2.1 \%$ (Table~\ref{tab:results}; Fig.~\ref{fig:ach}) on par with DreamerV3 ($14.5\pm1.6\%$; t-test $p=0.10$) and only outperformed by introducing prioritized experience replay ($19.4 \pm 1.6 \%$; \citealp{kauvar2023}).
To check that these results did indeed depend on the prediction of deterministic target embeddings, 
we trained the same model without  $\mathcal{L}_\mathrm{CDP}$, which is equivalent to classical Dreamer without $\mathcal{L}_\mathrm{recon}$. This manipulation resulted in an expected performance drop ($3.2\pm1.2\%$; Fig.~\ref{fig:abl1}).
We next compared \ours{} to other approaches. 
MuDreamer exhibited a notably lower Crafter score of $7.3\pm2.6\%$ (Table~\ref{tab:results}). 
This performance gap can likely be attributed to the comparatively weak action signal in Crafter. 
While we did not train DreamerPro, \citet{kauvar2023} reported a Crafter score of $4.7\pm0.5\%$. 
Thus, \ours{} performs on par with Dreamer and better than previous reconstruction-free approaches.

To check to what extent reward prediction contributed to \ours{}'s performance, we retrained the model without propagating gradients from the reward predictor head. 
We found that this ablation resulted in an intermediate performance drop to $12.7\pm1.6\%$ (Fig.~\ref{fig:abl1}).
In contrast, when training \ours{} without the alignment objectives $\mathcal{L}_\mathrm{dyn/rep}$, performance decreased to $6.3\pm1.9\%$.
Thus \ac{CDP} is necessary but not sufficient to improve reconstruction-free world models. 

\begin{table}[h]
    \centering
    \small
    \begin{tabular}{lccccc}
        \toprule
        Metrics &  Dreamer & DreamerPro & MuDreamer & \ours{} (ours)\\
        \midrule
        Crafter score & $14.5 \pm 1.6\%^\dagger$ &  $4.7 \pm 0.5\%^\dagger$ &  $7.3 \pm 2.6\%$ &  $16.2 \pm 2.1\%$ \\
        Cum.\ reward &  $11.7 \pm 1.9^\dagger$ & --- &  $5.6 \pm 1.6$ &  $9.8 \pm 0.4$ \\
        \bottomrule
    \end{tabular}
    \caption{ Crafter score and cum.\ reward $\pm$\,std ($n=7$) of different Dreamer variants (cf.\ Sec.~\ref{baseline_methods}). $\dagger$~published results by \citet{hafner2023} and \citet{kauvar2023}.} 
    \label{tab:results}
\end{table}

\section{Conclusion}

In this work, we showed that including \ac{CDP} is essential for learning a reconstruction-free world model that matches the  reconstruction-based Dreamer reference implementation on Crafter. 
An important direction for future research is to identify and benchmark other environments in which predictive learning provides advantages. 
On the one hand, we expect computational savings owed to removing the decoder in complex environments. 
On the other hand, we believe that reconstruction-free world models open the door to improved data-efficiency in complex high-dimensional environments with simple action signals and sparse reward structure. 

\subsubsection*{Acknowledgments}
The authors thank Ashena Gorgan Mohammadi, Peter Buttaroni, Fabian Mikulasch, and all members of the Zenke Lab for their thoughtful input.
This project was supported by the Swiss National Science Foundation (Grant Number PCEFP3\_202981), EU’s Horizon Europe Research
and Innovation Programme (Grant Agreement No. 101070374, CONVOLVE) funded through SERI (Ref. 1131–
52302), and the Novartis Research Foundation.

\bibliography{iclr2026_conference}
\bibliographystyle{iclr2026_conference}

\newpage
\appendix
\renewcommand\thefigure{\thesection.\arabic{figure}} 
\section{Appendix}   
\setcounter{figure}{0}    
\subsection{Hyperparameters}
\label{appendix:hyperparams}
We used the default parameters of the DreamerV3 XL model. $\beta_\mathrm{CDP} = 500$. The predictor was a two-layer MLP with 8192 input units, 4096 hidden units, and 4096 output units. The training ratio was 32 instead of 512. To ensure stable learning dynamics, the learning rate of the RSSM, and the predictor was trained with a higher learning rate of $4\cdot10^{-4}$ compared to the encoder ($\mathrm{lr}=6\cdot10^{-6}$). All other networks were trained with a learning rate of $4\cdot10^{-5}$.

\subsection{Supplementary Figures}

\begin{figure}[tbh]
    \centering
    \includegraphics[width=\linewidth]{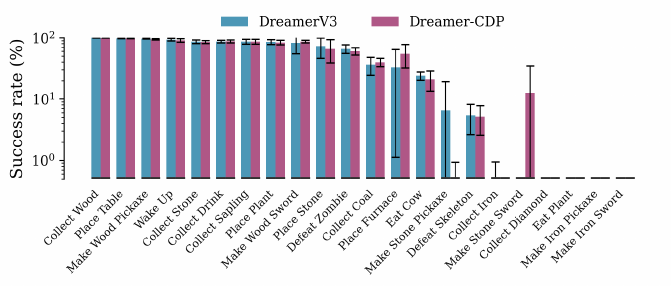}
    \caption{Achievement success rate for DreamerV3 (Blue) and \ours{} (purple) sorted by DreamerV3 success rates.}
    \label{fig:ach}
\end{figure}

\begin{figure}[tbh]
    \centering
    \includegraphics[width=\linewidth]{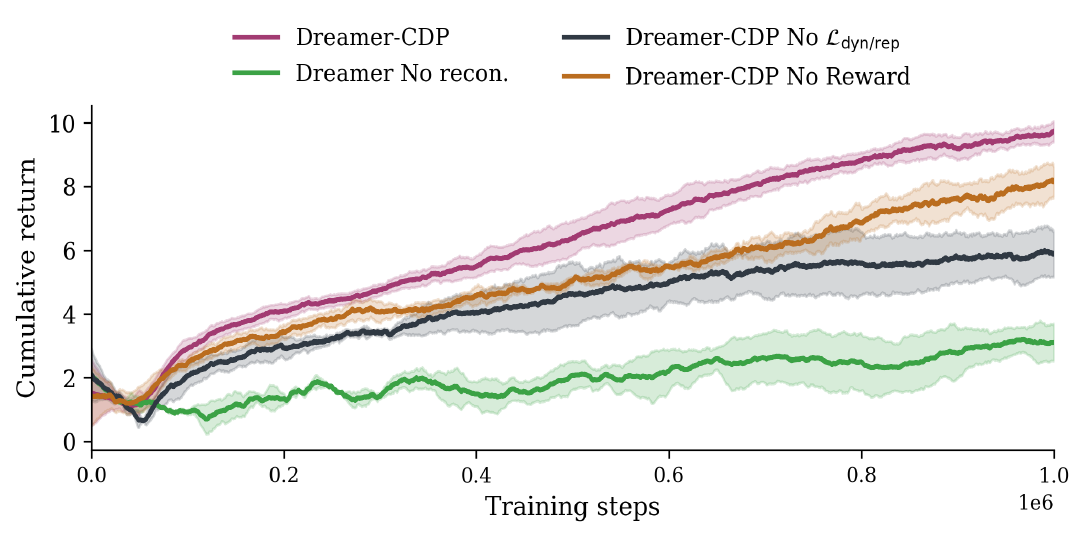}
    \caption{Comparison between \ours{} (Purple) and several ablations. Orange:~Ablation of the reward and value gradient to train the world model. Most of the learning signal comes from the latent space predictive loss. Green:~Ablation of $\mathcal{L}_\mathrm{CDP}$ and $\mathcal{L}_\mathrm{dyn/rep}$. The \ac{KL} balancing is not sufficient to train the world model in the latent space. Black:~CDP loss alone also results in lower cumulative return. }
    \label{fig:abl1}
\end{figure}

\end{document}